\documentclass[10pt,twocolumn,letterpaper]{article}
\usepackage[utf8]{inputenc}
\usepackage{iccv}
\usepackage{times}
\usepackage{epsfig}
\usepackage{graphicx}
\usepackage{amsmath}
\usepackage{amssymb}
\usepackage{mynotation}
\usepackage{algorithm}
\usepackage[noend]{algpseudocode}
\algrenewcommand{\algorithmicrequire}{\textbf{Input:}}
\algrenewcommand{\algorithmicensure}{\textbf{Output:}}

\newcommand{\algcomment}[1]{\hspace{3mm}{\footnotesize\# #1}}

\usepackage{makecell}
\usepackage{booktabs}
\usepackage[symbol]{footmisc}

\graphicspath{{./figures/}}

\usepackage[pagebackref=true,breaklinks=true,letterpaper=true,colorlinks,bookmarks=false]{hyperref}

\iccvfinalcopy 

\begin{document}

\title{Discriminative Online Learning for Fast Video Object Segmentation}

\author{Andreas Robinson\footnotemark[1] \footnotemark[2]\and
	Felix Järemo Lawin\footnotemark[1] \footnotemark[2]
	\and
	Martin Danelljan\footnotemark[2] \footnotemark[3]
	\and
	Fahad Shahbaz Khan\footnotemark[2] \footnotemark[4]
	\and
	Michael Felsberg\footnotemark[2]}
\maketitle

\footnotetext[1]{Authors contributed equally}
\footnotetext[2]{Computer Vision Laboratory, Linköping University, Sweden}
\footnotetext[3]{Computer Vision Laboratory, ETH Zürich, Switzerland}
\footnotetext[4]{Inception Institute of Artificial Intelligence, Abu Dhabi, UAE}

\begin{abstract}

We address the highly challenging problem of video object segmentation. Given only the initial mask, the task is to segment the target in the subsequent frames. In order to effectively handle appearance changes and similar background objects, a robust representation of the target is required. Previous approaches either rely on fine-tuning a segmentation network on the first frame, or employ generative appearance models. Although partially successful, these methods often suffer from impractically low frame rates or unsatisfactory robustness.

We propose a novel approach, based on a dedicated target appearance model that is exclusively learned online to discriminate between the target and background image regions. Importantly, we design a specialized loss and customized optimization techniques to enable highly efficient online training. Our light-weight target model is integrated into a carefully designed segmentation network, trained offline to enhance the predictions generated by the target model. Extensive experiments are performed on three datasets. Our approach achieves an overall score of over $70$ on YouTube-VOS, while operating at 25 frames per second.

\end{abstract}

\begin{figure}
	\centering
	\includegraphics[width=0.99\columnwidth]{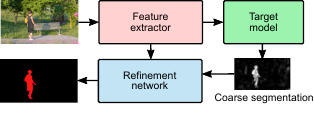}
	\includegraphics[width=0.49\columnwidth]{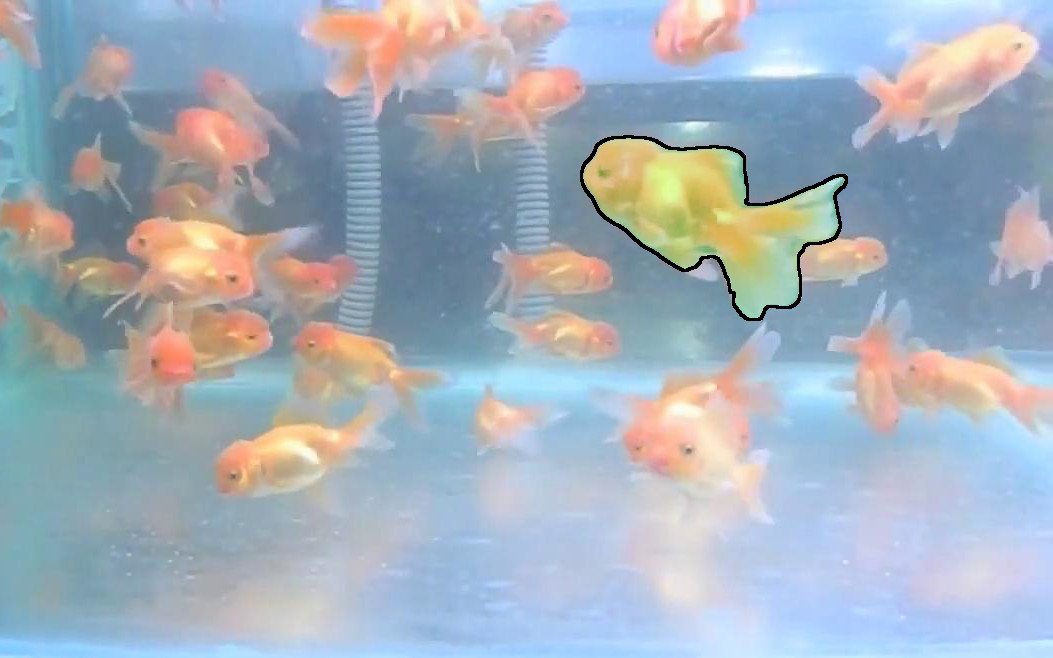}
	\includegraphics[width=0.49\columnwidth]{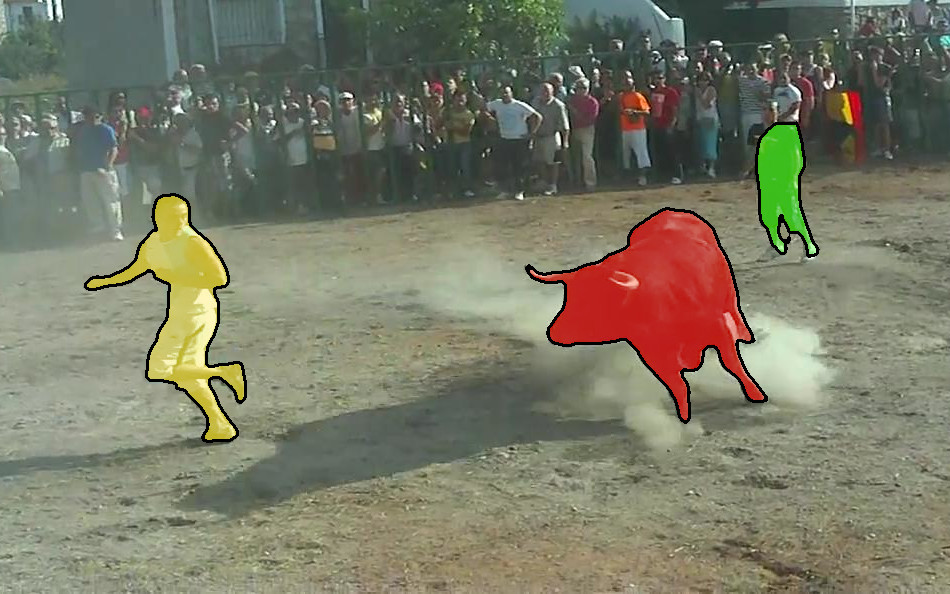}
	\caption{In our video segmentation approach (top), we first extract image features using a pre-trained ResNet-101. These features are then processed by our target appearance model, which is trained online to generate a robust coarse segmentation of the target. Guided by the image features, these segmentation scores are enhanced and upsampled by the refinement network. Our approach demonstrates significant robustness in the presence of background distractors (bottom left) and appearance changes (bottom right), owing to the discriminative and adaptive capabilities of the proposed target model.}
	\label{fig:intro}
	\vspace{-2mm}
\end{figure}

\section{Introduction}

Video object segmentation is one of the fundamental problems within the field of computer vision, with numerous applications in robotics, surveillance, autonomous driving and action recognition. The task is to predict pixel-accurate masks of the region occupied by a specific target object, in every frame of a given video sequence. In this work we focus on the semi-supervised setting, where the ground truth mask of the target is provided in the first frame. Challenges arise in dynamic environments with similar background objects and when the target undergoes considerable appearance changes or occlusions. Successful video object segmentation therefore requires both robust and accurate pixel classification of the target region. 

In the pursuit of robustness, several methods choose to fine-tune a segmentation network on the first frame, given the ground-truth mask. Although this has the potential to generate accurate segmentation masks under favorable circumstances, these methods suffer from extremely low frame-rates. Moreover, such extensive fine-tuning is prone to overfit to a single view of the scene, while degrading generic segmentation functionality learned during offline training. 
This limits performance in more challenging videos involving drastic appearance changes, occlusions and distractor objects in the background \cite{xu2018youtube2}.

In this work, we propose a novel approach to address the challenges involved in video object segmentation. We introduce a dedicated target appearance model that is exclusively learned online, to discriminate the target region from the background. This model is integrated into our final segmentation architecture by designing a refinement network that produces high-quality segmentations based on predictions generated by the target model.

Unlike the methods relying on first-frame fine-tuning, our refinement network is trained offline to be target agnostic. Consequently, it retains powerful and generic object segmentation functionality, while target-specific learning is performed entirely by the target appearance module. A simplified illustration of our network is provided in figure \ref{fig:intro}.

We propose a discriminative online target model, capable of robustly differentiating between the target and background, by addressing the following challenges: (i) The model must be inferred solely from online data. (ii) The learning procedure needs to be highly efficient in order to maintain low computational complexity. (iii) The model should be capable of robust and controlled online updates to handle extensive appearance changes and distractor objects. All these aspects are accommodated by learning a light-weight fully-convolutional network head, with the goal to provide coarse segmentation scores of the target. We design a suitable objective and utilize specialized optimization techniques in order to efficiently learn a highly discriminative target model during online operation. 

With the aim of constructing a simple framework, we refrain from using optical flow, post-processing and other additional components.
Our final approach, consisting of a single deep network for segmentation and the target appearance model, is easily trained on video segmentation data in a single phase.
Comprehensive experiments are performed on three benchmarks: DAVIS 2016 \cite{perazzi2016davis}, DAVIS 2017 \cite{perazzi2016davis} and YouTube-VOS \cite{xu2018youtube2}. We first analyze our approach in a series of ablative comparisons and then compare it to several state-of-the-art approaches. Our method achieves an overall score of 73.4 on DAVIS 2017 and 71.0 on YouTube-VOS, while operating at 15 frames per second. In addition, a faster version of our approach achieves 25 frames per second, with only slight degradation in segmentation accuracy.

\section{Related work}

The task of video object segmentation has seen extensive study and rapid development in recent years, largely driven by the introduction and evolution of benchmarks such as DAVIS \cite{perazzi2016davis} and YouTube-VOS \cite{xu2018youtube}.

\noindent\textbf{First-frame fine-tuning:} Most state-of-the-art approaches train a segmentation network offline, and then fine-tune it on the first frame \cite{perazzi2017masktrack, caelles2017osvos, maninis2017osvos_s, xu2018youtube2} to learn the target-specific appearance. This philosophy was extended \cite{voigtlaender2017onavos} by additionally fine-tuning on subsequent video frames. Other approaches \cite{cheng2017segflow, hu2018mgcrn, luiten2018premvos} further integrate optical flow as an additional cue. While obtaining impressive results on the DAVIS 2016 dataset, the extensive fine-tuning leads to impractically long running times. Furthermore, such extensive fine-tuning is prone to over-fitting, a problem only partially addressed by heavy data augmentation \cite{khoreva2017lucid}.

\noindent\textbf{Non-causal methods:}
Another line of research tackles the video object segmentation problem by allowing non-causal processing, by e.g.\ fine-tuning over blocks of video frames \cite{li2018dyenet}. Other methods \cite{bao2018cinm, jang2017ctn} adopt spatio-temporal Markov Random Fields (MRFs), or infer discriminative location specific embeddings \cite{ci2018video}. In this work, we focus on the \emph{causal} setting in order to accommodate real-time applications. 

\noindent\textbf{Mask propagation:} Several recent methods \cite{perazzi2017masktrack,oh2018rgmp, yang2018osnm,johnander2018generative} employ a mask-propagation module to improve spatio-temporal consistency of the segmentation. In \cite{perazzi2017masktrack}, the model is learned offline to predict the target mask through refinement of the previous frame's segmentation output. RGMP~\cite{oh2018rgmp} attempts to further avoid first-frame fine-tuning by concatenating the current frame features with a target representation generated in the first frame. A slightly different approach is proposed in \cite{yang2018osnm}, where the mask of the previous frame is represented as a spatial Gaussian and a modulation vector. Unlike these methods, we are not relying on spatio-temporal consistence assumptions imposed by the mask-propagation approach. Instead, we use previous segmentation masks to train the discriminative model.

\noindent\textbf{Generative approaches:} Another group of methods \cite{behl2018meta,vondrick2018tracking,chen2018blazingly,hu2018videomatch,voigtlaender2018feelvos} incorporate light-weight generative models of the target object. Rather than fine-tuning the network on the first frame, these methods first construct appearance models from features corresponding to the initial target labels. Features from incoming frames are then evaluated by the appearance model with techniques inspired by classical machine learning clustering methods \cite{chen2018blazingly,johnander2018generative} or feature matching approaches \cite{hu2018videomatch,voigtlaender2018feelvos}.

\noindent\textbf{Tracking:} Visual object tracking is similar to video object segmentation in that both problems involve following a specific object; although the former outputs bounding boxes rather than a pixel-wise segmentation. The problem of efficient online learning of discriminative target-specific appearance models has been extensively explored in visual tracking \cite{henriques2015high,hare2016struck}. Recently, optimization-based trackers, such as ECO \cite{danelljan2017eco} and ATOM \cite{danelljan2018atom}, have achieved impressive results on benchmarks. These methods train convolution filters to discriminate between target and background.

The close relation between the two problem domains is made explicit in video object segmentation methods such as \cite{cheng2018favos}, where object trackers are used as external components to locate the target. In contrast, we do not employ off-the-shelf trackers to predict the target location. We instead take inspiration from optimization-based learning employed in recently introduced trackers \cite{danelljan2017eco, danelljan2018atom}, to train a discriminative target model online, combining it with a refinement network trained offline.

\newcommand{\bI}{{\bf I}}  
\newcommand{\bK}{\mathcal{M}}  
\newcommand{\bx}{{\bf x}}  
\newcommand{\bs}{{\bf s}}  
\newcommand{\by}{{\bf y}}  
\newcommand{\bw}{{\bf w}}  
\newcommand{\bv}{{\bf v}}  

\section{Method}

\begin{figure*}[h]
\includegraphics[width=\textwidth]{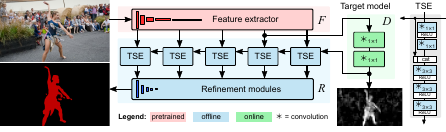}
\caption{Overview of our video segmentation architecture, consisting of the feature extractor $F$, the refinement network $R$ and target appearance model $D$. The pre-trained feature extractor $F$ first produces feature maps from the input image (top left), at five depths. The online-trained target model $D$ then generates coarse target segmentation scores (bottom right) from the second-deepest feature map. The scores and all five feature maps are finally passed to the offline-trained refinement network $R$, to produce the output segmentation mask (bottom left). $R$ consists of the refinement modules and target segmentation encoder (TSE) blocks (detailed in the gray inset, right).}
\label{fig:system}
\end{figure*}

In this work we propose a method for video object segmentation, integrating a powerful target appearance model into a deep neural network. The target model is trained online on the given sequence to differentiate between target and background appearance. We employ a specialized optimization technique, which is crucial for efficiently training the target model online and in real-time. All other parts of the network is fully trained offline, thereby avoiding any online fine-tuning.

Our approach has three main building blocks: a feature-extraction network $F$, a target model $D$ and a refinement network $R$. The target model $D$ generates a coarse prediction of the target location given deep features as input. These predictions are then improved and up-sampled by $R$, which is trained offline. We train the target model $D$ online on a dataset $\bK$ with target labels $\by_i$ and corresponding features $\bx_i$. The full architecture is illustrated in figure \ref{fig:system}.

During inference, we train $D$ on the first frame with the given ground-truth target mask and features extracted by $F$. In the subsequent frames, the target model first predicts a coarse target segmentation score map $\bs$. This score map together with features from $F$ are processed by $R$, which outputs a high quality segmentation mask. The mask and associated features are stored in $\bK$ and are used to update $D$ before the next incoming frame.

\subsection{Target model}
\label{sec:discriminator}

We aim to develop a powerful and discriminative model of the target appearance, capable of differentiating between the target and background image regions. To successfully accommodate the video object segmentation problem, the model must be robust to significant appearance changes and distractor objects. Moreover, it needs to be easily updated with new data and efficiently trainable. Since these aspects have successfully been addressed in visual tracking, we take inspiration from recent research in this related field in the development of our approach. We employ a light-weight discriminative model that is exclusively trained online. Our target model $D(\bx; \bw)$, parameterized by $\bw$, takes an input feature map $\bx$ and outputs coarse segmentation scores $\bs$. 
For our purpose, the target model $D$ is realized as two fully-convolutional layers,
\begin{equation}
D({\bf x};{\bf w}) = {\bf w}_2 * ({\bf w}_1 * {\bf x}) \,.
\end{equation}
We use a factorized formulation for efficiency, where the initial layer $\bw_1$ reduces the feature dimensionality while the second layer $\bw_2$ computes the segmentation scores.

The target model $D$ is trained online based on image features $\bx$ and the corresponding target segmentation masks $\by$. Fundamental to our approach, the target model parameters $\bw$ need to be learned with minimal computational impact. To enable the deployment of specific fast converging optimization techniques we adopt an $L^2$ loss. Our online learning loss is given by,
\begin{equation}
\mathcal{L}_{D}({\bf w}; \bK) = 
\sum_k\gamma_k
\left\Vert \bv_k \! \cdot \! (\by_k - U(D(\bx_k)) \right\Vert^2 + \sum_j \lambda_j \left\Vert {\bf w}_j \right\Vert^2 \!,
\label{eq:L2}
\end{equation}
Here, the parameters $\lambda_j$ control the regularization term and $\bv_k$ are weight masks balancing the impact of target and background pixels. $U$ denotes bilinear up-sampling of the output from the target model to the spatial resolution of the labels $\by_k$. The dataset memory $\bK = \{(\bx_k, \by_k, \gamma_k)\}_{k=1}^{K}$ consists of sample feature maps $\bx_k$, target labels $\by_k$ and sample weights $\gamma_k$. During inference, $\bK$ can easily be updated with new samples from the video sequence. Compared to blindly fine-tuning on the latest frame, the dataset $\bK$ provides a controlled way of adding new data while keeping past frames in memory by setting appropriate sample weights $\gamma_k$.

\noindent\textbf{Training:}
To minimize the target model loss over $\bw$ in equation \eqref{eq:L2} we employ the Gauss-Newton (GN) optimization strategy proposed in \cite{danelljan2018atom}. This is an iterative method, requiring an initial $\bw$. In each iteration the optimal increment $\Delta \bw$ is found using a quadratic approximation of the loss
\begin{equation}
\mathcal{L}_{D}(\bw +\Delta \bw) \approx \Delta \bw^T J_{\bw}^T J_{\bw}\Delta\bw + 2 \Delta \bw^T J_{\bw}^T r_{\bw} + r_{\bw}^T r_{\bw}\,.
\label{eq:gn}
\end{equation}
Here $J_{\bw}$ is the Jacobian of $\mathcal{L}_{D}$ at $\bw$ and $r_{\bw}$ contain the residuals $\sqrt{\gamma_k} \bv_k \cdot (\by_k - U(D(\bx_k)))$ and $\sqrt{\lambda_j} \bw_j$. The loss in \eqref{eq:gn} results in a positive definite quadratic problem, which is minimized over $\Delta \bw$ with Conjugate Gradient (CG) descent \cite{hestenes1952methods}. We then update $\bw \leftarrow \bw+ \Delta \bw $ and execute the next GN iteration. 

\noindent\textbf{Pixel weighting:}
To address the imbalance between target and background, we employ a weight mask $\bv$ in \eqref{eq:L2} to ensure that the target influence is not too small relative to the usually much larger background region. We define the target influence as the fraction of target pixels in the image, or $\hat{\kappa}_k = N^{-1}\sum_{n} (\by_k)_{n}$ where $n$ is the pixel index and $N$ the total number of pixels. The weight mask is then defined as
\begin{equation}
{\bf v}_k = 
\begin{cases}
\kappa/\hat{\kappa}_k, & ({\bf y}_k)_{n} = 1 \\
(1-\kappa) / (1-\hat{\kappa}_k), & ({\bf y}_k)_{n} = 0
\end{cases}
\end{equation}
where $\kappa = \min(\kappa_\mathrm{min}, \hat{\kappa}_k)$ is the desired and $\hat{\kappa}_k$ the actual target influence. We set $\kappa_\mathrm{min} = 0.1$ in our approach.

\noindent\textbf{Initial sample generation:}
In the first frame of the sequence, we train the target model on the initial dataset $\bK_0$, created from the given target mask $\by_0$ and the extracted features $\bx_0$. To add more variety in the initial frame, we generate additional augmented samples. Based on the initial label $\by_0$, we first cut out the target object and apply a fast inpainting method \cite{telea2004inpainting} to restore the background. We then apply a random affine warp and blur before pasting the target object back onto the image, creating a set of augmented images $\tilde{\bI}_k$ and corresponding label masks $\tilde{\by}_k$. 
After a feature extraction step, the unmodified first frame and the augmented frames are combined into the dataset $\bK_0 = \{(\tilde{\bx}_k, \tilde{\by}_k, \gamma_k)\}_{k=0}^{K-1}$.
We initialize the sample weights $\gamma_k$ such that the original sample carries twice the weight of the other samples, due to its higher importance. Finally, the weights are normalized to sum to one.

\subsection{Refinement Network}
\label{sec:refinement}
While the target model provides robust but coarse segmentation scores, the final aim is to generate an accurate segmentation mask of the target at the original image resolution. To this end, we introduce a refinement network, that processes the coarse score $\bs$ along with backbone features. The network consists of two types of building blocks: a target segmentation encoder (TSE) and a refinement module.
From these we construct a U-Net based architecture for object segmentation as in \cite{yu2018dfn}. Unlike most state-of-the-art methods for semantic segmentation \cite{zhao2017pyramid, chen2018deeplab}, the U-Net structure does not rely on dilated convolutions, but effectively integrates low-resolution deep feature maps. This is crucial for reducing the computational complexity of our target model during inference. 

The refinement network takes features maps $\bx^d$ from multiple depths in the backbone network as input. The resolution is decreased by a factor of 2 at each depth $d$. For each layer, the features $\bx^d$ along with the coarse scores $\bs$ are first processed by the corresponding TSE block $T^d$. The refinement module $R^d$ then inputs the resulting segmentation encoding generated by $T^d$ and the refined outputs ${\bf z}^{d+1}$ from the preceding deeper layer, 
\begin{equation}
{\bf z}^d = R^d(T^d(\bx^{d}, \bs) , {\bf z}^{d+1})\,.
\end{equation}
The refinement modules are comprised of two residual blocks and a channel attention block, as in \cite{yu2018dfn}. For the deepest block we set ${\bf z}^{d+1}$ to an intermediate projection of $\bx^d$ inside $T^d$. The output ${\bf z}^1$ at the shallowest layer is processed by a residual block providing the final refined segmentation output $\hat{\by}$. 

\noindent\textbf{Target segmentation encoder:} Seeking to integrate features and scores, we introduce the target segmentation encoder (TSE). It processes features in two steps, as visualized in figure \ref{fig:system} (gray inset, right). We first project the backbone features to 64 channels to reduce the subsequent computational complexity. Additionally, we maintain 64 channels throughout the refinement network, keeping the number of parameters low. After projection, the features are concatenated with the segmentation score $\bs$ and encoded by three convolutional layers.

\subsection{Offline training}
\label{sec:offline}
During the offline training phase, we learn the parameters $\theta_R$ in our refinement network on segmentation data. The network is trained on samples consisting of one reference frame and one or more validation frames. These are all randomly selected from the same video sequence.
A training iteration is performed as follows: We first learn the target model weights $\bw$, described in section \ref{sec:discriminator}, based on the reference frame only. We then apply our full network, along with the learned target model, on the validation frames to predict the target segmentations. The parameters in the network are trained by minimizing the binary cross-entropy loss with respect to the ground-truth masks.

The backbone network is a ResNet-101, pre-trained on ImageNet. During offline training, we only learn the parameters of the refinement module, and freeze the weights of the backbone.
Since the target model only receives backbone features, we can pre-learn and store the target model weights for each sequence. The offline training time is therefore  not significantly affected by the learning of $D$. 

The network is trained on a combination of the YouTube-VOS~\cite{xu2018youtube2} and DAVIS 2017~\cite{perazzi2016davis} train splits. These are balanced such that DAVIS 2017 is traversed eight times per epoch, and YouTube-VOS once. We select one reference frame and two validation frames per sample and train refinement network with the ADAM optimizer \cite{kingma2014adam}. 

We start with the learning rate $\alpha=10^{-3}$ and moment decay rates $\beta_1=0.9, \beta_2=0.999$ with weight decay $10^{-5}$, and train for 120 epochs. The learning rate is then reduced to $\alpha=10^{-4}$, and we train for another 60 epochs.

\subsection{Inference}
\label{sec:inference}

\begin{algorithm}[t]
	\caption{Inference}
	\begin{algorithmic}[1]
		\Require Images $\bI_i$, target $\by_0$
		\State $i=0$, $\bx_0 = F(\bI_0)$, 
		\State$\bK_0 = \{(\tilde{\bx}_k, \tilde{\by}_k, \gamma_k)\}_{k=1}^{K}$ \algcomment{initialize dataset}
		\State ${\bf w_0} = \mathrm{optimize}( \mathcal{L}_D(\bw;\bK_0)$) \algcomment{Init $D$, sec \ref{sec:discriminator}}
		\For{ $i = 1, 2, \ldots$ }
		\State	$\bx_i = F(\bI_i)$ \algcomment{Extract features}
		\State	$\bs_i = D(\bx_i; \bw_{i-1})$ \algcomment{Predict target, sec \ref{sec:discriminator}}
		\State	${\bf \hat y}_i = R(\bx_i, \bs_i; \theta_R)$ \algcomment{Refine target, sec \ref{sec:refinement}}
		\State	$\bK_i = \mathrm{extend}(\bx_i, \hat{\by}_i\, \gamma_i;\bK_{i-1})$ \algcomment{extend dataset}
		\If{$i\mod{t_s} = 0$} \algcomment{Update $D$ every $t_s$ frame}
		\State ${\bf w}_{i} = \mathrm{optimize}( \mathcal{L}_D(\bw;, \bK_i))$ 
		\EndIf
		\EndFor
	\end{algorithmic}
	\label{alg:deploy}
\end{algorithm}

During inference, we apply our video segmentation procedure, also summarized in algorithm \ref{alg:deploy}. We first generate the augmentation dataset $\bK_0$, as described in \ref{sec:discriminator}, given the initial image $\bI_0$ and the corresponding target mask $\by_0$. 

We then optimize the target model $D$ on this dataset. In the consecutive frames, we first predict the coarse segmentation scores $\bs$ using $D$. Next, we refine $\bs$ with the network $R$ (section \ref{sec:refinement}). The resulting segmentation output $\hat{\by}_i$ along with the input features $\bx_i$ are added to the dataset. Each new sample $(\bx_i, \hat{\by}_i, \gamma_i)$ is first given a weight $\gamma_i = (1-\eta)^{-1}\gamma_{i-1}$, with $\gamma_0 = \eta$. We then normalize the sample weights to sum to unity. The parameter $\eta<1$ controls the update rate, such that the most recent samples in the sequence are prioritized in the re-optimization of the target model $D$. For practical purposes, we limit the maximum capacity $K_{\mathrm{max}}$ of the dataset. When the maximum capacity is reached, we remove the sample with the smallest weight from $\bK_{i-1}$ before inserting a new one. In all our experiments we set $\eta = 0.1$ and $K_{\mathrm{max}}=80$.

During inference, we optimize the target model parameters $\bw_{1}$ and $\bw_{2}$ on the current dataset $\bK_{i}$ every $t_s$-th frame. For efficiency, we keep the first layer of the target model $\bw_{1}$ fixed during updates. Setting $t_s$ to a large value reduces the inference time and regularizes the update of the target model. On the other hand, it is important that the target model is updated frequently, for objects that undergo rapid appearance changes. In our approach we set $t_s=8$, i.e.\ we update $D$ every eighth frame.
 
The algorithm can be trivially expanded into handling multiple object segmentation. We then employ one online target model for each object and fuse the final refined predictions with softmax aggregation as proposed in \cite{oh2018rgmp}. Note that we only require one feature extraction per frame, since the image features $\bx_i$ are common for all target objects.

\subsection{Implementation details}
\label{sec:imp-details}

We implement our method in the PyTorch framework \cite{paszke2017automatic} and use its pre-trained ResNet-101 as the basis of the feature extractor $F$. Following the naming convention in table 1 in \cite{He2015}, we extract five feature maps from the max pooling output in "conv2\_x" and the outputs of blocks "conv2\_x" through "conv5\_x". The target model $D$ accepts 1024-channel features from "conv4\_x" and produces 1-channel score maps. Both the input features and output scores have a spatial resolution 1/16th of the input image.

\noindent\textbf{Target model:} The first layer $\bw_{1}$ has $1 \times 1$ kernels reducing input features to $c=96$ channels while $\bw_{2}$ has a 3 $\times$ 3 kernel with one output channel. During first-frame optimization, $\bw_{1}$ and $\bw_{2}$ are randomly initialized by Kaiming Normal~\cite{he2015delving}. Using the data augmentation in \ref{sec:discriminator}, we generate a initial dataset $\bK_0$ of 20 image and label pairs. We then optimize $\bw_{1}$ and $\bw_{2}$ with the Gauss-Newton algorithm outlined in section \ref{sec:discriminator} in $N_\text{GN}=5$ GN steps. We apply $N_\text{CGi}=5$ CG iterations in the first step and $N_\text{CG}=10$ in the others. As the starting solution is randomly initialized we typically use fewer CG iterations in the first GN step. In the target model update step we use $N_\text{CGu}=10$ CG iterations, updating $\bw_{2}$ every $t_s=8$ frame, while keeping $\bw_{1}$ fixed. We employ the aforementioned settings in our final approach, denoted {\bf Ours} in the following sections.

The architecture allows us to alter the settings of the target model without retraining the refinement network. Taking advantage of this, we additionally develop a fast version, termed \textbf{Ours (fast)} by reducing the number of optimization steps, the number of filters in $\bw_{1}$ and increasing the update interval $t_s$. Specifically, we set $c=32$, $t_s=16$, $N_\text{GN}=4$, $N_\text{CGi}=5$, $N_\text{CG}=10$ and $N_\text{CGu}=5$.

\section{Experiments}

\newcommand{\mcJ}{\mathcal{J}}  
\newcommand{\mcF}{\mathcal{F}}  
\newcommand{\mcG}{\mathcal{G}} 
\newcommand{\mcJF}{\mathcal{J \& F}} 
We perform experiments on three benchmarks: DAVIS 2016 \cite{perazzi2016davis}, DAVIS 2017 \cite{perazzi2016davis} and YouTube-VOS \cite{xu2018youtube2}. For YouTube-VOS, we compare on the official test-dev set, with withheld ground-truth. For ablative experiments, we also show results on a separate validation split of the YouTube-VOS train set, consisting of 300 videos not used for training. Following the standard DAVIS protocol, we report both the mean Jaccard $\mcJ$ index and mean boundary $\mcF$ scores, along with the overall score $\mcJF$, which is the mean of the two. For comparisons on YouTube-VOS, we report $\mcJ$ and $\mcF$ scores for classes included in the training set (seen) and the ones that are not (unseen). The overall score $\mcG$ is computed as the average over all four scores, defined in YouTube-VOS. 
In addition, we compare the computational speed of the methods in terms of frames per second (fps), computed by taking the average over the DAVIS 2016 validation set. For our approach, we computed frame-rates on a single GPU. Further results are provided in the supplement.

\subsection{Ablation}  

We analyze the contribution of all key components in our approach. All compared approaches are trained using the same data and settings.

\noindent\textbf{Base net:} First, we construct a baseline network to analyze the impact of our discriminative appearence model $D$. This is performed by replacing $D$ with an offline-trained target encoder, while retraining the refinement network $R$. As for our proposed network we keep the backbone $F$ parameters fixed. The target encoder is comprised of two convolutional layers, taking reference frame features from $conv4\_x$ and the corresponding target mask as input. Features ($conv4\_x$) extracted from the test frame are concatenated with the output from the target encoder and processed with two additional convolutional layers. The output is then passed to the refinement network $R$ in the same manner as for the coarse segmentation score $\bs$ (see section~\ref{sec:refinement}). We train this model with the same methodology as for our proposed network. 

\noindent\textbf{Base net + F.-T:} To compare our discriminative online learning with first-frame fine-tuning of the network, we further integrate a fine-tuning strategy into the Base net above. For this purpose, we create an initial dataset $\bK_0$ with 20 samples using the \emph{same} sample generation procedure employed for our approach (section~\ref{sec:discriminator}). We then fine-tune all components in the architecture (refinement network $R$ and the target-encoder), except the backbone feature extractor. For fine-tuning we use the ADAM optimizer with 100 iterations and a batch size of four.

\noindent\textbf{$D$-only:} To analyze the impact of the refinement network $R$, we simply remove it from our architecture and instead let the target-specific model $D$ output the final segmentations. The target model's coarse predictions are upsampled to full image resolution through bilinear interpolation. We learn the segmentation threshold by training the scale and offset parameters prior to the output sigmoid operation. In this version, we only train the target model $D$ on the first frame, and refrain from subsequent updates.

\noindent\textbf{Ours - no update:} To provide a fair comparison with the aforementioned versions, we evaluate a variant of our approach that does not perform any update of the target model $D$ during inference. The model $D$ is thus only trained on the first frame of the video.

\noindent\textbf{Ours:} Finally, we enable target model updates (as described in section \ref{sec:inference}) to obtain our final approach.

In table \ref{tab:ab}, we present the results in terms of $\mcJ$ score on the separate validation split of the YouTube-VOS dataset. The base network, not employing the target model $D$ achieves a score of $52.0\%$. Adding fine-tuning on the first frame leads to an absolute improvement of $7.2\%$. Using only the target model $D$ outperforms online fine-tuning. This is particularly notable, considering that the $D$-only version contains only two offline-trained parameters in the network. Our target model thus possesses powerful discriminative capabilities, achieving high robustness. Further adding the refinement network $R$ (Ours - no update) leads to a major absolute gain of $8.8\%$. This improvement stems from the offline-learned processing of the coarse segmentations, yielding more accurate mask predictions. Finally, the proposed online updating strategy additionally improves the score to $70.6\%$. 

\newcommand{\yes}{\checkmark}
\begin{table}[t!]
	\centering
	\resizebox{0.8\columnwidth}{!}{%
	\begin{tabular}[]{lccccc}
		\toprule
		Version & $D$ & $R$ & Update & F.-T. & $\mcJ$ \\
		\midrule
		Base net & & \yes & & & 52.0 \\
		Base net + F.-T. & & \yes & & \yes & 57.2 \\
		$D$-only & \yes & & & &58.8  \\
		\textbf{Ours} - no update & \yes & \yes &   & & 67.6  \\
		\textbf{Ours} & \yes & \yes & \yes & & \textbf{70.6}   \\
		\bottomrule
	\end{tabular}}\vspace{1mm}
	\caption{Ablative study on a validation split of 300 sequences from YouTube-VOS. We analyze the different components of our approach where $D$ and $R$ denote the target model and refinement module respectively. Further, "Update" indicates if the target model update is enabled and "F.-T." denotes first-frame fine-tuning. Our target appearance model $D$ outperforms the Base net methods. Further, the refinement network significantly improves the raw predictions from the target model $D$. Finally, the best performance is obtained when additionally updating target model $D$.}
	\label{tab:ab}
	\vspace{-2mm}
\end{table}

\subsection{Comparison to state-of-the-art}

We compare our method to a variety of recent approaches on the YouTube-VOS, DAVIS 2017 and DAVIS 2016 benchmarks. We provide results for two versions of our approach: \textbf{Ours} and \textbf{Ours (fast)} (see section~\ref{sec:imp-details}).

\noindent\textbf{YouTube-VOS~\cite{xu2018youtube2}}: The official YouTube-VOS test-dev dataset has 474 sequences with objects from 91 classes. Out of these, 26 classes are not present in the training set. We compare our method with the results reported in \cite{xu2018youtube}, that were obtained by retraining the methods on the YouTube-VOS training set. Additionally, we compare to the recent AGAME \cite{johnander2018generative} method, also trained on YouTube-VOS.

The results are reported in table~\ref{tab:ytvos_results}. OSVOS~\cite{caelles2017osvos}, OnAVOS~\cite{voigtlaender2017onavos} and S2S~\cite{xu2018youtube} all employ first-frame fine-tuning, obtaining inferior frame-rates below $0.3$ FPS. Among these methods, the recent S2S achieves the best overall performance with a $\mathcal{G}$-score of $64.4$. The AGAME method, employing a generative appearance model and no fine-tuning, obtains a $\mathcal{G}$-score of $66.0$. Our approach significantly outperforms all previous methods by a relative margin of $7.6\%$, yielding a final $\mathcal{G}$-score of $71.0$. Moreover, our approaches achieve the fastest inference speeds. Notably, Ours (fast) maintains an impressive $\mathcal{G}$-score of $70.3$, with an average segmentation speed of $25.3$ FPS.

\begin{table}[t!]
\centering
\resizebox{0.85\columnwidth}{!}{%
\centering
\begin{tabular}{lcccc}
\toprule
       & $\mathcal{G}$ & $\mathcal{J}$ & $\mathcal{F}$ & FPS\\
Method & \footnotesize overall & \footnotesize seen $|$ unseen & \footnotesize seen $|$ unseen &    \\
\midrule
\textbf{Ours}  & \textbf{71.0} & \textbf{71.6} $|$ \textbf{65.0} & \textbf{74.7} $|$ 72.5 & 14.6\\
\textbf{Ours} (fast)  & 70.3 & 69.9 $|$ 64.9 & 73.2 $|$ \textbf{73.1} & \textbf{25.3}\\
\midrule
AGAME \cite{johnander2018generative} & 66.0 & 66.9 $|$ 61.2 & 68.6 $|$ 67.3 & 14.3\\
S2S \cite{xu2018youtube} & 64.4 & 71.0 $|$ 55.5 & 70.0 $|$ 61.2 & 0.11\\
OnAVOS \cite{voigtlaender2017onavos} & 55.2 & 60.1 $|$ 46.1 & 62.7 $|$ 51.4 & 0.08\\
OSVOS \cite{caelles2017osvos} & 58.8 & 59.8 $|$ 54.2 & 60.5 $|$ 60.7 & 0.22\\
\bottomrule
\end{tabular}

}\vspace{1mm}
\caption{State-of-the-art comparison on the large-scale YouTube-VOS test-dev dataset, containing 474 videos. The results of our approach were obtained through the official evaluation server. We report the mean Jaccard ($\mcJ$) and boundary ($\mcF$) scores for object classes that are \emph{seen} and \emph{unseen} in the training set, along with the overall mean ($\mathcal{G}$). Our approaches achieve superior performance, with $\mathcal{G}$-scores over $70$, while operating at high frame-rates.}
\label{tab:ytvos_results}
\vspace{-2mm}
\end{table}

\begin{table}[t]
	\centering
	\tabcolsep=0.1cm
	\resizebox{\columnwidth}{!}{%
	\newcommand{\first}[1]{\textbf{\textcolor{red}{#1}}}
\newcommand{\second}[1]{\textit{\textcolor{blue}{#1}}}
\begin{tabular}{@{~~~~}l@{~~~~}ccc@{~~~~}|@{~~~~}ccc@{~~~~}|@{~~~~}c@{~~~~}}  
          \toprule
          & \multicolumn{3}{c@{~~~~~~~}}{DAVIS17}  & \multicolumn{3}{c@{~~~~~~}}{DAVIS16} &  \\
          Method & $\mcJF$ & $\mathcal{J}$  & $\mathcal{F}$  & $\mcJF$ & $\mathcal{J}$  & $\mathcal{F}$ & FPS\\
\midrule
\textbf{Ours}  & \second{73.4} & \second{71.3} & \second{75.5} & 84.8 & 85.0 & 84.5 & \second{14.6} \\
\textbf{Ours} (fast)  & 70.9 & 68.4 & 73.3 & 82.5 & 82.7 & 82.4 & \first{25.3} \\
\textbf{Ours} (DV17)  & 67.5 & 65.8 & 69.2 & 83.2&  82.7 & 83.7 & \second{14.6} \\
\midrule
AGAME \cite{johnander2018generative} & 70.0 & 67.2 & 72.7 & - & 82.0 & - & 14.3 \\
RGMP \cite{oh2018rgmp} & 66.7 & 64.8 & 68.6 & 81.8 & 81.5 & 82.0 & 7.7 \\
FEELVOS \cite{voigtlaender2018feelvos} & 71.5 & 69.1 & 74.0 & 81.7 & 81.1 & 82.2 & 2.22 \\
VM \cite{hu2018videomatch} & - & 56.6 & - & - & 81.0 & - & 8.33 \\
FAVOS \cite{cheng2018favos} & 58.2 & 54.6 & 61.8 & 80.8 & 82.4 & 79.5 & 0.56 \\
OSNM \cite{yang2018osnm} & 54.8 & 52.5 & 57.1 & - & 74.0 & - & 7.14 \\
PReMVOS \cite{luiten2018premvos} & \first{77.8} & \first{73.9} & \first{81.7} & \first{86.8} & 84.9 & \first{88.6} & 0.03 \\
OSVOS-S \cite{maninis2017osvos_s} & 68.0 & 64.7 & 71.3 & \second{86.5} & \second{85.6} & \second{87.5} & 0.22 \\
OnAVOS \cite{voigtlaender2017onavos} & 67.9 & 64.5 & 71.2 & 85.5 & \first{86.1} & 84.9 & 0.08 \\
MGCRN \cite{hu2018mgcrn} & - & - & - & 85.1 & 84.4 & 85.7 & 1.37 \\
CINM \cite{bao2018cinm} & 67.5 & 64.5 & 70.5 & - & 84.2 & - & 0.00 \\
\bottomrule
\end{tabular}

	}\vspace{1mm}
	\caption{Comparison our method with current state-of-the art approaches on the DAVIS 2017 and DAVIS 2016 validation sets.  The best and second best entries are shown in red and blue respectively. In addition to Ours and Ours (fast), we report the results of our approach when trained on \emph{only} DAVIS 2017, in Ours (DV17). Our segmentation method outperform compared methods with practical frame-rates. Furthermore, we achieve competitive results even when trained with only DAVIS 2017, owing to the discriminative nature of our target model.}
	\vspace{-2mm}
	\label{tab:dvjoint_results}
\end{table}

\noindent\textbf{DAVIS 2017~\cite{perazzi2016davis}}: The validation set for DAVIS 2017 contains 30 sequences. In addition to Ours and Ours (fast), we now include a third version, Ours (DV17). To create this version, we train a separate refinement network employing \emph{only} the DAVIS 2017 train split. No other datasets, such as PASCAL VOC, is employed. The target model in this version has the same parameters as Ours, (see section \ref{sec:imp-details}). 

We report the results on DAVIS 2017 in table~\ref{tab:dvjoint_results}. The methods OnAVOS~\cite{voigtlaender2017onavos}, OSVOS-S~\cite{maninis2017osvos_s}, MGCRN~\cite{hu2018mgcrn} and CINM~\cite{bao2018cinm} employ extensive fine-tuning on the first-frame, experiencing impractical segmentation speed. FAVOS~\cite{cheng2018favos} utilizes part-based trackers and a segmentation network to track and segment the target. The methods VM~\cite{hu2018videomatch}, RGMP~\cite{oh2018rgmp}, OSNM~\cite{yang2018osnm}, AGAME~\cite{johnander2018generative} and FEELVOS~\cite{voigtlaender2018feelvos} all employ mask-propagation. In addition, AGAME, VM and FEELVOS utilize generative appearance models. Our approach outperforms all aforementioned methods with a $\mathcal{J \& F}$ average of $73.4$, while simultaneously reaching the highest frame-rate.

When training on only DAVIS 2017 data, Ours (DV17), achieves an $\mathcal{J \& F}$ average of $67.5$. This is comparable to the performance of OnAVOS~\cite{voigtlaender2017onavos}, OSVOS-S~\cite{maninis2017osvos_s} and RGMP~\cite{oh2018rgmp}, which utilize much additional data during training, such as PASCAL VOC~\cite{Everingham15} and MS COCO~\cite{lin2014microsoft}. This demonstrates that our refinement network can be trained robustly with limited data, and illustrates the superior generalization capabilities of our target model.

Among the compared methods, PReMVOS~\cite{luiten2018premvos} was not discussed above, since it constitutes a major framework, encompassing multiple components and cues: extensive fine-tuning, mask-region proposals, optical flow based mask predictions, multiple refinement steps, merging and tracking modules, and re-identification. While this approach obtains an $\mathcal{J \& F}$ average of $77.8$, it is approximately 500 times slower than ours. In contrast, our approach is simple, consisting of a single network together with a light-weight target model. However, many of the components in PReMVOS are complementary to our work, and can be directly integrated. Our aim is to propose a simple and general framework to stimulate future research.

\noindent\textbf{DAVIS 2016~\cite{perazzi2016davis}}: Finally, we evaluate our method on the 20 validation sequences in DAVIS 2016, corresponding to a subset of DAVIS 2017 and report the results in table~\ref{tab:dvjoint_results}. Our method performs comparable to the fine-tuning based approaches PReMVOS~\cite{luiten2018premvos}, MGCRN~\cite{hu2018mgcrn}, OnAVOS~\cite{voigtlaender2017onavos} and OSVOS-S~\cite{maninis2017osvos_s}. Further, all three versions of our method outperform AGAME~\cite{johnander2018generative}, RGMP~\cite{oh2018rgmp}, OSNM~\cite{yang2018osnm}, FAVOS~\cite{cheng2018favos}, VM~\cite{hu2018videomatch} and FEELVOS~\cite{voigtlaender2018feelvos}. 

\begin{figure*}[!t]
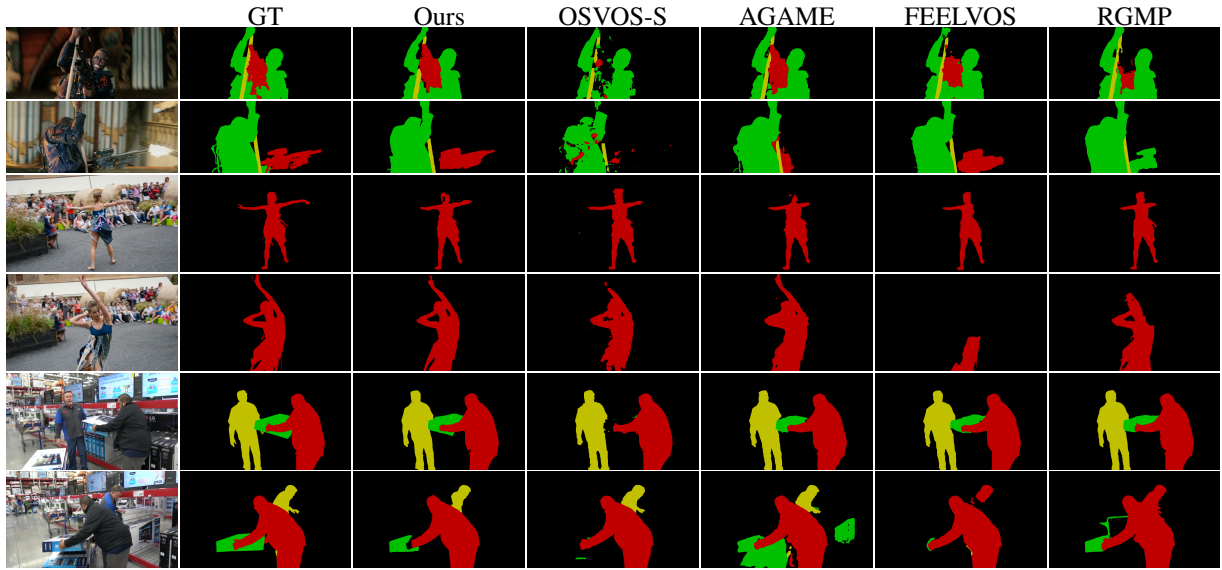

	\centering 
	\newcommand{\image}{\includegraphics[width=0.13\textwidth]}
	\tabcolsep=0.02cm
	\renewcommand{\arraystretch}{0.3}
	\begin{tabular}{ccccccc}
 & GT & Ours & OSVOS-S & AGAME & FEELVOS & RGMP \\

\image{qresults/shooting/00004.jpg} &
\image{qresults/shooting/00004-gt.png} &
\image{qresults/shooting/00004.png} &
\image{qresults/shooting/00004-osvoss.png} &
\image{qresults/shooting/00004-agame.png} &
\image{qresults/shooting/00004-feelvos.png} &
\image{qresults/shooting/00004-rgmp.png} \\
\image{qresults/shooting/00039.jpg} &
\image{qresults/shooting/00039-gt.png} &
\image{qresults/shooting/00039.png} &
\image{qresults/shooting/00039-osvoss.png} &
\image{qresults/shooting/00039-agame.png} &
\image{qresults/shooting/00039-feelvos.png} &
\image{qresults/shooting/00039-rgmp.png} \\

\image{qresults/dance-twirl/00030.jpg} &
\image{qresults/dance-twirl/00030-gt.png} &
\image{qresults/dance-twirl/00030.png} &
\image{qresults/dance-twirl/00030-osvoss.png} &
\image{qresults/dance-twirl/00030-agame.png} &
\image{qresults/dance-twirl/00030-feelvos.png} &
\image{qresults/dance-twirl/00030-rgmp.png} \\
\image{qresults/dance-twirl/00089.jpg} &
\image{qresults/dance-twirl/00089-gt.png} &
\image{qresults/dance-twirl/00089.png} &
\image{qresults/dance-twirl/00089-osvoss.png} &
\image{qresults/dance-twirl/00089-agame.png} &
\image{qresults/dance-twirl/00089-feelvos.png} &
\image{qresults/dance-twirl/00089-rgmp.png} \\

\image{qresults/loading/00005.jpg} &
\image{qresults/loading/00005-gt.png} &
\image{qresults/loading/00005.png} &
\image{qresults/loading/00005-osvoss.png} &
\image{qresults/loading/00005-agame.png} &
\image{qresults/loading/00005-feelvos.png} &
\image{qresults/loading/00005-rgmp.png} \\
\image{qresults/loading/00049.jpg} &
\image{qresults/loading/00049-gt.png} &
\image{qresults/loading/00049.png} &
\image{qresults/loading/00049-osvoss.png} &
\image{qresults/loading/00049-agame.png} &
\image{qresults/loading/00049-feelvos.png} &
\image{qresults/loading/00049-rgmp.png} \\
\end{tabular}

	\vspace{1mm}
	\caption{Examples of three sequences from DAVIS 2017, demonstrating how our method performs under significant appearance changes compared to ground truth, OSVOS-S \cite{maninis2017osvos_s}, AGAME \cite{johnander2018generative}, FEELVOS \cite{voigtlaender2018feelvos} and RGMP \cite{oh2018rgmp}.}
	\label{fig:qualitative_results2}
	\vspace{-2mm}
\end{figure*}
\begin{figure}
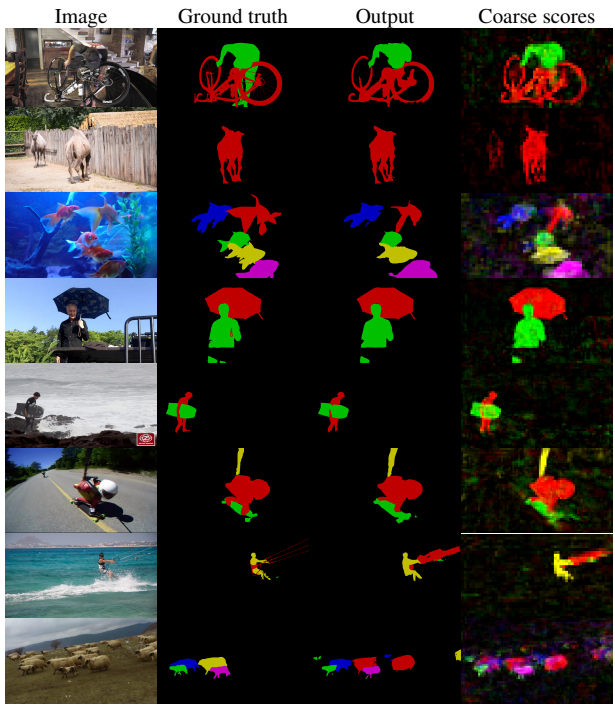

\newcommand{\image}{\includegraphics[width=0.24\columnwidth]}
\centering 
\tabcolsep=0.01cm
\renewcommand{\arraystretch}{0.06}
\begin{tabular}{cccc}

{\footnotesize Image} & {\footnotesize Ground truth} & {\footnotesize Output} & {\footnotesize Coarse scores} \\

\image{qresults/bike-packing/00034.jpg} &
\image{qresults/bike-packing/00034-gt.png} &
\image{qresults/bike-packing/00034.png} &
\image{qresults/bike-packing/00034-scores.png} \\

\image{qresults/camel/00089.jpg} &
\image{qresults/camel/00089-gt.png} &
\image{qresults/camel/00089.png} &
\image{qresults/camel/00089-scores.png} \\

\image{qresults/gold-fish/00050.jpg} &
\image{qresults/gold-fish/00050-gt.png} &
\image{qresults/gold-fish/00050.png} &
\image{qresults/gold-fish/00050-scores.png} \\


\image{qresults/14dae0dc93/00050.jpg} &
\image{qresults/14dae0dc93/00050-gt.png} &
\image{qresults/14dae0dc93/00050.png} &
\image{qresults/14dae0dc93/00050-scores.png} \\

\image{qresults/731b825695/00020.jpg} &
\image{qresults/731b825695/00020-gt.png} &
\image{qresults/731b825695/00020.png} &
\image{qresults/731b825695/00020-scores.png} \\

\image{qresults/4f414dd6e7/00175.jpg} &
\image{qresults/4f414dd6e7/00175-gt.png} &
\image{qresults/4f414dd6e7/00175.png} &
\image{qresults/4f414dd6e7/00175-scores.png} \\


\image{qresults/kite-surf/00040.jpg} &
\image{qresults/kite-surf/00040-gt.png} &
\image{qresults/kite-surf/00040.png} &
\image{qresults/kite-surf/00040-scores.png} \\

\image{qresults/a9c9c1517e/00045.jpg} &
\image{qresults/a9c9c1517e/00045-gt.png} &
\image{qresults/a9c9c1517e/00045.png} &
\image{qresults/a9c9c1517e/00045-scores.png} \\

\end{tabular}
\vspace{1mm}
\caption{Qualitative results of our method, including both segmentation masks and target model score maps. The top three examples shown are from the DAVIS 2017 validation set, followed by three examples from our YouTube-VOS validation set. The last two examples are representative failure cases from both datasets.}

\label{fig:qualitative_results1}
\vspace{-3mm}
\end{figure}
\subsection{Qualitative Analysis}
\noindent\textbf{Target model:}
In figure~\ref{fig:qualitative_results1}, we visualize the segmentation scores provided by our target model along with the final segmentation output. We analyze the output on sequences from the DAVIS 2017 and Youtube-VOS validation sets. For the first six rows in figure~\ref{fig:qualitative_results1}, our approach is able to accurately segment the target object. The target model (right) provide a discriminative coarse segmentation scores, that are robust to distractor objects, even in the most challenging cases (third row). Moreover, our refinement network successfully enhances the already accurate predictions from the target model, generating final segmentation masks with impressive level of detail. Additionally, the refinement network has the ability to suppress erroneous predictions generated by the target model, e.g the second camel in the second row, adding further robustness.

The last two rows in figure \ref{fig:qualitative_results1} demonstrate two cases, where our method struggles. Firstly, in the kite surfing sequence, our method detects the thin kite lines properly, but fails to separate them from the background. Accurately segmenting such thin structures in the image is a highly challenging task. Secondly, in the bottom row, our target model has failed to distinguish between instances of sheep due to their very similar appearance. 

\noindent\textbf{State-of-the-art:}
In figure \ref{fig:qualitative_results2} we visually compare our approach on three challenging sequences, with state-of-the-art methods employing fine-tuning (OSVOS-S \cite{maninis2017osvos_s}), mask-propagation (RGMP \cite{oh2018rgmp}) with generative modeling (AGAME \cite{johnander2018generative}) and feature matching (FEELVOS \cite{voigtlaender2018feelvos}). 

In the first sequence (first and second row) the extensively fine-tuned network in OSVOS-S fails to generalize to the changing target poses. While the mask-propagation in RGMP and generative appearance model in AGAME are more accurate, they both fail to segment the red target (a weapon). Further, the occlusions imposed by the twirling dancer in the middle sequence (third and fourth row), are causing the feature matching approach in FEELVOS to lose track of the target. Finally, in the last sequence (fifth and sixth row), all of the above methods fail in robustly segmenting the three targets. In contrast, our method accurately segments all targets in all of the selected sequences, demonstrating the robustness of our appearance model.

\section{Conclusion}

We present a novel approach to video object segmentation integrating a light-weight but highly discriminative target appearance model with a segmentation network. The discriminator produces coarse but robust target predictions and the segmentation network subsequently refine these into high-quality target segmentation masks. The target appearance model is efficiently trained online while the completely target-agnostic segmentation network is trained offline. Our method achieves state-of-the-art performance on the YouTube-VOS dataset and competitive results on DAVIS 2017, while operating at frame rates superior to previous methods.

{\small
\bibliographystyle{ieee}
\bibliography{references}
}

\end{document}